\documentclass{ecai}
\usepackage{times}
\usepackage{graphicx}
\usepackage{latexsym}

\usepackage{graphicx}
\usepackage{float} 
\usepackage{subfigure}
\usepackage{amsmath}
\usepackage{amsfonts}
\usepackage{algorithm}      
\usepackage[noend]{algpseudocode} 
\usepackage{wrapfig}    
\usepackage{amssymb}
\usepackage{mathtools}

\begin{document}

\title{Combinatorial Optimization by Graph Pointer Networks and Hierarchical Reinforcement Learning}

\author{Qiang Ma\institute{Columbia University, Department of Computer Science, email: ma.qiang@columbia.edu} \qquad Suwen Ge \qquad Danyang He \qquad Darshan Thaker \qquad Iddo Drori\institute{Columbia University, Department of Computer Science, email: idrori@cs.columbia.edu and Cornell University, School of Operations Research and Information Engineering, email: idrori@cornell.edu}\\ \\ Columbia University}
\maketitle
\bibliographystyle{ecai}

\begin{abstract}
In this work, we introduce Graph Pointer Networks (GPNs) trained using reinforcement learning (RL) for tackling the traveling salesman problem (TSP). GPNs build upon Pointer Networks by introducing a graph embedding layer on the input, which captures relationships between nodes. Furthermore, to approximate solutions to constrained combinatorial optimization problems such as the TSP with time windows, we train hierarchical GPNs (HGPNs) using RL, which learns a hierarchical policy to find an optimal city permutation under constraints. Each layer of the hierarchy is designed with a separate reward function, resulting in stable training. Our results demonstrate that GPNs trained on small-scale TSP50/100 problems generalize well to larger-scale TSP500/1000 problems, with shorter tour lengths and faster computational times. We verify that for constrained TSP problems such as the TSP with time windows, the feasible solutions found via hierarchical RL training outperform previous baselines. In the spirit of reproducible research we make our data, models, and code publicly available. 
\end{abstract}


\section{INTRODUCTION}
%
%
%
\noindent As a fundamental problem in Computer Science and Operations Research, combinatorial optimization problems have received wide attention in the past few decades. One of the most important and practical problems is the traveling salesman problem (TSP). To introduce the TSP, consider a salesman who is traveling on a tour across a set of cities. The salesman must visit all cities exactly once while minimizing the overall tour length. TSP is known to be an NP-complete problem \cite{papadimitriou1977euclidean}, which captures the difficulty of finding efficient exact solutions in polynomial time. To overcome this complexity barrier, several approximation algorithms and heuristics have been proposed such as the 2-opt heuristic \cite{aarts2003local}, Christofides algorithm \cite{christofides1976worst}, guided local search \cite{voudouris1999guided}, and the Lin-Kernighan heuristic (LKH) \cite{helsgaun2000effective}.

With the development of machine learning (ML) and reinforcement learning (RL), an increasing number of recent works concentrate on solving combinatorial optimization using an ML or RL approach \cite{vinyals2015pointer,bello2016neural,nazari2018reinforcement,li2018combinatorial,khalil2017learning,kool2018attention,kool2019buy,joshi2019efficient}. A \textit{seq2seq} model, known as the \textit{pointer network} \cite{vinyals2015pointer}, has great potential in approximating solutions to several combinatorial optimization problems such as finding the convex hull and the TSP. It uses LSTMs as the encoder and an attention mechanism \cite{vaswani2017attention} as the decoder to extract features from city coordinates. It then predicts a policy that describes the next possible move so that a permutation of visited cities is sampled. An RL framework for pointer networks has been proposed \cite{bello2016neural}, in which the pointer network model is trained by the Actor-Critic algorithm \cite{mnih2016asynchronous} and the negative tour length is used as a reward. The RL approach proved to be more efficient than previous supervised learning methods and outperformed most of the previous heuristics on TSP with up to 100 nodes. As an extension of the pointer network, Nazari et al. \cite{nazari2018reinforcement} modified the architecture of the pointer network to tackle more complex combinatorial optimization problems, such as the vehicle routing problem (VRP). 

Due to the property of routing problems, the neural network architectures used in the above works do not fully take into account the relationship between problem entities, which is a critical property of routing problems and also plays a role in several other problems. As a powerful tool to process non-Euclidean data and capture graph information, Graph Neural Networks (GNNs) \cite{kipf2016semi,xu2018powerful} have been studied extensively in recent years. 
Based on GNNs, two novel approaches \cite{li2018combinatorial,khalil2017learning} were proposed, which leverage the information of the inherent graph structure present in many combinatorial optimization problems. Li et al. \cite{li2018combinatorial} applied a Graph Convolutional Network (GCN) model \cite{kipf2016semi} along with a guided tree search algorithm to solve graph-based combinatorial optimization problems such as Maximal Independent Set and Minimum Vertex Cover problems. Dai et al. \cite{khalil2017learning} proposed a graph embedding network trained with deep Q-learning and found that this generalized well to larger-scale problems. Recently, motivated by the Transformer architecture \cite{vaswani2017attention}, Kool et al. proposed an attention model \cite{kool2018attention,kool2019buy} to solve routing problems such as the TSP, VRP, and Orienteering Problem. In their model, the relationships between the nodes of the graph are captured by a multi-head attention mechanism, using a rollout baseline in the REINFORCE algorithm, which significantly improves the result for small-scale TSP. However, scale is still an issue for the attention model.

The previous works have achieved good approximate results on various combinatorial optimization problems, but combinatorial optimization problems with constraints, e.g. TSP with time window (TSPTW), have not been fully considered. To deal with constrained problems, Bello et al. \cite{bello2016neural} proposed a penalty method, which added a penalty term for infeasible solutions on the reward function. However, the penalty method can lead to unstable training, and the hyperparameters of the penalty term are usually difficult to tune. A better choice for training is using \textit{hierarchical RL} methods, which have been applied widely to tackle complex problems such as video games with sparse rewards and robot maze tasks \cite{kulkarni2016hierarchical,nachum2018data,haarnoja2018latent}. The key motivation for hierarchical RL is the splitting of complex tasks into several simple subproblems which are learned in a hierarchy. Haarnoja et al. \cite{haarnoja2018latent} introduced latent space policies for hierarchical RL, in which the lower layers of the hierarchy provide a feasible solution space and constrain the actions of the higher layers. The higher layers then make decisions based on the information from the latent space in the lower layers. In this work, we explore the use of hierarchical RL methods to tackle combinatorial optimization problems with constraints, which are split into different subtasks. Each layer of the hierarchy learns to search the feasible solutions under constraints or learns the heuristics to optimize the objective function. 

In this work, we aim to approximate solutions to larger-scale TSP problems and address constrained combinatorial optimization problems. The contributions of this work are three-fold: Firstly, we propose a \textit{graph pointer network} (GPN) to tackle the vanilla TSP. The GPN extends the pointer network with graph embedding layers and achieves faster convergence. 
Secondly, we add a vector context to the GPN architecture and train using early stopping in order to generalize our model to tackle larger-scale TSP instances, e.g. TSP1000, from a model trained on a much smaller TSP50 instance. Thirdly, 
we employ a hierarchical RL framework along with the GPN architecture to efficiently solve TSP with a time window constraint. For each task, we conduct experiments to compare our model performance with existing baselines and previous work. 


This work is structured as follows. In the Preliminaries section, we formulate the TSP and its corresponding reinforcement learning framework. The Hierarchical Reinforcement Learning section introduces the hierarchical RL framework as well as the hierarchical policy gradient method. The Graph Pointer Network section describes the architecture of the proposed GPN and its hierarchical version. Then, in the Experiments section, we analyze our approach on small-scale TSP problems, their generalization capabilities to large-scale TSP problems, as well as their performance on the TSP with Time Windows problem.

\section{PRELIMINARIES} \label{sec:preliminaries}
\subsection{Traveling Salesman Problem}
In this work, we focus on solving the symmetric 2-D Euclidean traveling salesman problem (TSP) \cite{lawler1985traveling}. The graph of the symmetric TSP is complete and undirected. Given a list of $N$ city coordinates $\{\mathbf{x}_1,\mathbf{x}_2,...,\mathbf{x}_N\} \subset\mathbb{R}^2$, the problem is to find an optimal route such that each city is visited exactly once and the total distance covered in the route is minimized. In other words, we wish to find an optimal permutation $\sigma$ over the cities that minimizes the tour length \cite{bello2016neural}:
\begin{equation}
L(\sigma, \mathbf{X})= \sum_{i=1}^{N}\| \mathbf{x}_{\sigma(i)}-\mathbf{x}_{\sigma(i+1)} \|_2,
\end{equation}
where $\sigma(1)=\sigma(N+1)$, $\sigma(i)\in\{1,...,N\}$,
$\sigma(i)\neq\sigma(j)$ for any $i\neq j$, and $\mathbf{X}= [\mathbf{x}_1^{\top},...,\mathbf{x}_N^{\top}]^{\top}\in \mathbb{R}^{N \times 2}$ is a matrix consisting of all city coordinates $\mathbf{x}_i$.
In addition, in our work, we consider the TSP with added constraints. Generally, the constrained TSP is written as the following optimization problem:
\begin{equation}
\begin{array}{cl}
\min\limits_{\sigma}  & L(\sigma, \mathbf{X})= \sum\limits_{i=1}^{N}
 \|\mathbf{x}_{\sigma(i)}-\mathbf{x}_{\sigma(i+1)}\|_2\\
 \mbox{s.t.} &  f(\sigma,\mathbf{X})= 0,\\
 &  g(\sigma,\mathbf{X})\leq 0,
\end{array}
\end{equation}
where $\sigma$ is a permutation, $f(\sigma,\mathbf{X})$ and $g(\sigma,\mathbf{X})$ represent constraint functions.


\subsection{Reinforcement Learning for TSP}
We begin by introducing the notation used to formulate the TSP as a reinforcement learning problem. Let $\mathcal{S}$ be the state space and $\mathcal{A}$ be the action space. Each state $\mathbf{s}_t\in\mathcal{S}$ is defined as the set of all previous visited cities, i.e. $\mathbf{s}_t=\{\mathbf{x}_{\sigma(i)}\}_{i=1}^t$. 
The action $\mathbf{a}_t\in\mathcal{A}$ is defined as the next selected city, that is $\mathbf{a}_t=\mathbf{x}_{\sigma(t+1)}$. Since $\sigma(1)=\sigma(N+1)$, it follows that $\mathbf{a}_N=\mathbf{x}_{\sigma(N+1)}=\mathbf{x}_{\sigma(1)}$, which means the last choice of the route is the start city. 

Denote a policy as $\pi_{\theta}(\mathbf{a}_t|\mathbf{s}_t)$, which is a distribution over candidate cities $\mathbf{a}_t$ given a set of visited cities $\mathbf{s}_t$. Given a set of visited cities, the policy will return a probability distribution over the next candidate cities that have not been chosen. In our case, the policy is represented by a neural network and the parameter $\theta$ represents the trainable weights of the neural network. Furthermore, the reward function is defined as the negative cost incurred from taking action $a_t$ from state $s_t$, i.e. $r(\mathbf{s}_t,\mathbf{a}_t)=-\|\mathbf{x}_{\sigma(t)}-\mathbf{x}_{\sigma(t+1)}\|_2$. Then the expected reward \cite{sutton2018reinforcement} is defined as follows:
\begin{equation}\label{reward}
\begin{split}
&\mathbb{E}_{(\mathbf{s}_t,\mathbf{a}_t)\sim \pi_{\theta}(\mathbf{s}_t,\mathbf{a}_t)}\left[\sum_{i=1}^N
r(\mathbf{s}_t,\mathbf{a}_t)\right]
\\&= \mathbb{E}_{\sigma\sim p_\theta(\Gamma),\mathbf{X}\sim \mathcal{X}}\left[\sum_{i=1}^{N}-
\| \mathbf{x}_{\sigma(i)}-\mathbf{x}_{\sigma(i+1)} \|_2\right]\\
& = -\mathbb{E}_{\sigma\sim p_\theta(\Gamma),\mathbf{X}\sim \mathcal{X}}\left[L(\sigma,\mathbf{X})\right]
\end{split}
\end{equation}
where $\mathcal{X}$ is the space of the set of cities, $\Gamma$ is the space of all possible permutations $\sigma$ over $\mathcal{X}$, and $p_\theta(\Gamma)$ is the distribution over $\Gamma$, which is predicted by the neural network. To maximize the above reward function, the network must learn a policy to minimize the expected tour length. We employ the policy gradient algorithm \cite{sutton2018reinforcement} to learn to maximize the reward function as described next.


\section{HIERARCHICAL REINFORCEMENT LEARNING}
\label{sec:hrl}
\subsection{Hierarchical RL for TSP}
A key aspect of our work is tackling TSP with constraints. Augmenting traditional RL reward functions with a penalty term encourages solutions to be in the feasible set \cite{bello2016neural}; however, we find this method leads to unstable training. Instead, we propose a hierarchical RL framework to more efficiently tackle TSP with constraints. 

\begin{figure}[H]
\centering
\subfigure[Middle Layer]{
\label{GraphModel-middle}
\includegraphics[width=0.14\textwidth]{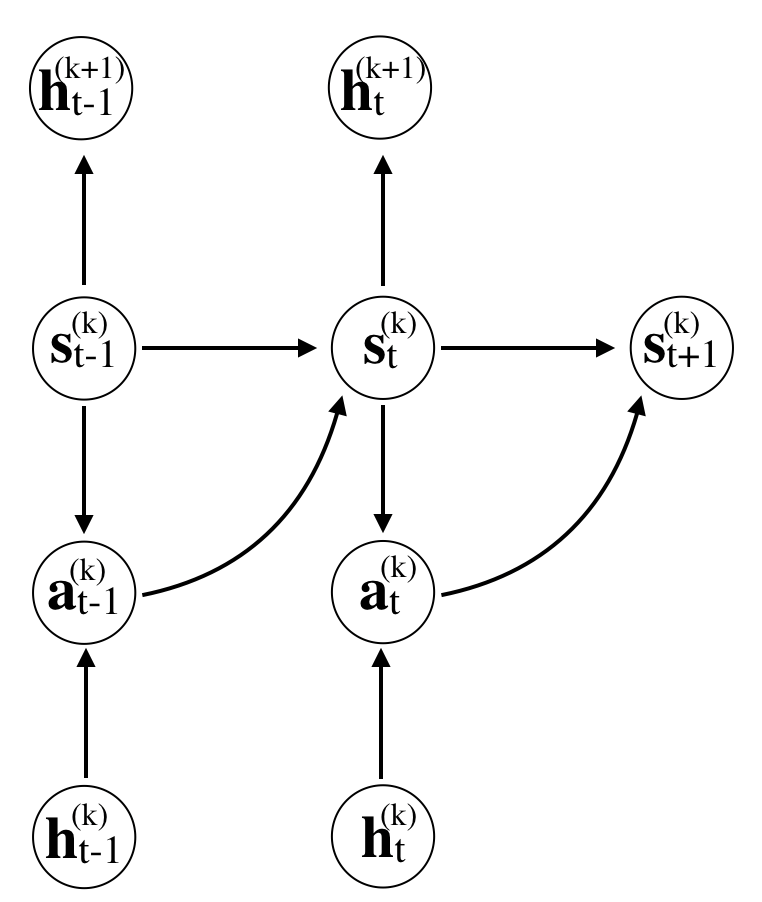}}
\subfigure[Lowest Layer]{
\label{GraphModel-bottom}
\includegraphics[width=0.14\textwidth]{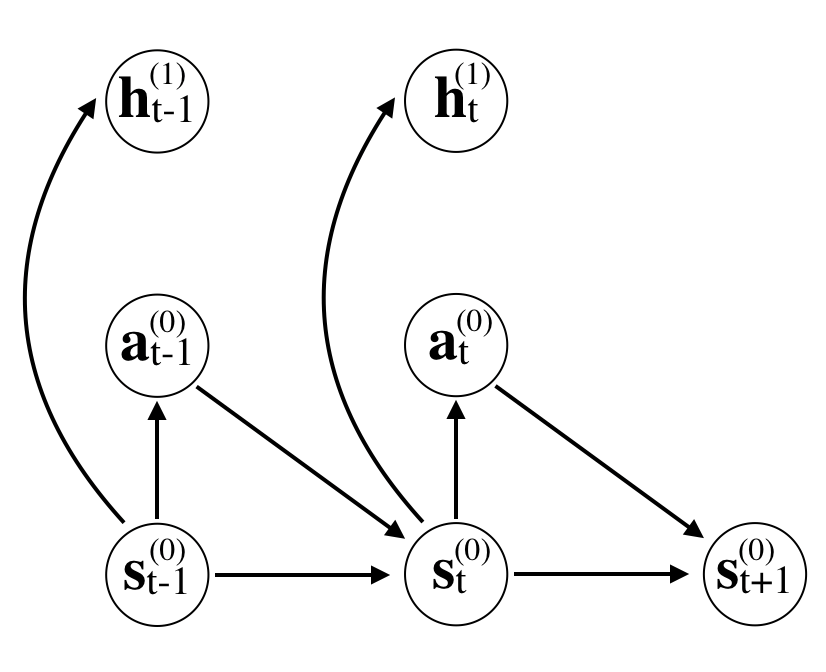}}
\subfigure[Highest Layer]{
\label{GraphModel-top}
\includegraphics[width=0.14\textwidth]{graphicalmodel_bottom.png}}
\caption{Graphical models for hierarchical RL framework. (a) Middle layer of hierarchy: In each middle layer, the next action is conditioned both on the current state and the latent variable from the lower layer. It also provides the latent variable for the next higher layer. (b) Lowest layer of hierarchy: a simple MDP which provides latent variables for the next layer. (c) Highest layer: does not provide latent variables and only utilizes latent variable from the lower layer.}
\label{GraphModel}
\end{figure}

Motivated by the work of Haarnoja et al. \cite{haarnoja2018latent,haarnoja2018soft}, we adopt a probabilistic graphical model framework for control, as demonstrated in Figure \ref{GraphModel}. Each layer of a hierarchy defines a policy, from which we sample actions. At a given layer $k \in \{0, \dots, K\}$, the current action $\mathbf{a}^{(k)}_{t}$ is sampled from the policy $\pi_{\theta_k}(\mathbf{a}^{(k)}_{t}|\mathbf{s}^{(k)}_{t},\mathbf{h}^{(k)}_{t})$, where $\mathbf{h}^{(k)}_{t}\in\mathcal{H}^{(k)}$ is a latent variable from the previous layer in the hierarchy and $\mathcal{H}^{(k)}$ is its corresponding latent space. The lowest layer shown in Figure \ref{GraphModel-bottom} is a simple Markov Decision Process (MDP) with action $\mathbf{a}_t^{(0)}$ sampled from policy $\pi_{\theta_0}(\mathbf{a}_{t}^{(0)}|\mathbf{s}_{t}^{(0)})$, which provides a latent vector $\mathbf{h}_{t}^{(1)}$ for the higher layer. The middle layer not only depends on the latent variable $\mathbf{h}_{t}^{(k)}$ from the $(k-1)$-th layer, but also provides a latent variable $\mathbf{h}_{t}^{(k+1)}$ for the next higher layer. For convenience of notation, on the $k$-th layer, we extend the policy to both sample the action and provide the latent variable, i.e.  $\mathbf{a}^{(k)}_{t},\mathbf{h}_{t}^{(k+1)}\sim\pi_{\theta_k}(\cdot|\mathbf{s}^{(k)}_{t},\mathbf{h}^{(k)}_{t})$.


Each layer corresponds to a different RL task, so the reward functions are hand-designed to be different for each layer. There are two natural ways to formulate constrained TSP optimization problems in a hierarchical fashion. First, we set lower layer reward functions to simply bias solutions to be in the feasible set of the constrained optimization problem, and set higher layer reward functions to be the original optimization objective. Conversely, we order reward functions in increasing difficulty of optimization: the first layer attempts to solve vanilla TSP, the second layer is given a TSP instance with one constraint, and so on. For our experiments, we use the first formulation, since we find that this yields better results.


\subsection{Hierarchical Policy Gradient}
We use the policy gradient method to learn a hierarchical policy. Considering a hierarchical policy, the objective function of the $k$-th layer is $ J(\theta_k)= -\mathbb{E}_{\sigma\sim p_{\theta_k(\sigma)},\mathbf{X}\sim \mathcal{X}}\left[L(\sigma,\mathbf{X})\right]$. Based on the REINFORCE algorithm, the gradient of the $k$-th layer policy is expressed as \cite{bello2016neural,williams1992simple}: 
\begin{equation}\label{hier-gradient}
\begin{split}
\nabla_{\theta_k}J(\theta_k)&=\frac{1}{B}\sum_{i=1}^B \Bigg[\left(\sum_{t=1}^N r_k(\mathbf{s}_{i,t}^{(k)},\mathbf{a}_{i,t}^{(k)}) -b_{i,k}\right)\\& \times \left(\sum_{t=1}^N\nabla_{\theta_k}\log\pi_{\theta_k}(\mathbf{a}_{i,t}^{(k)}|\mathbf{s}_{i,t}^{(k)}, \mathbf{h}_{i,t}^{(k)})\right)\Bigg],\\
\end{split}
\end{equation}
where $B$ is the batch size, $\pi_{\theta_k}$ is the $k$-th layer policy, $r_k(\cdot,\cdot)$ is the reward function for the $k$-th layer, $b_{i,k}$ is the $k$-th layer baseline, and $\mathbf{h}_{t}^{(k)}$ is the latent variable from the lower layer. 
Based on Equation \ref{hier-gradient}, the parameters $\theta_k$ are optimized using gradient descent through the update rule $\theta_k\leftarrow\theta_k+\alpha\nabla_{\theta_k}J(\theta_k)$. 
\subsubsection{Central Self-Critic}
We introduce the \textit{central self-critic} baseline $b_{i,k}$, which is similar to the self-critic baseline \cite{rennie2017self} and the rollout baseline in the Attention Model \cite{kool2018attention}. The central self-critic baseline $b_{i,k}$ is expressed as:
\begin{equation}\label{eq:central-self-critic}
\begin{split}
b_{i,k} &= \sum_{t=1}^N\left(r_k(\tilde{\mathbf{s}}_{i,t}^{(k)},\tilde{\mathbf{a}}_{i,t}^{(k)})\right) \\& + 
\left[\frac{1}{B}\sum_{t=1}^N \sum_{j=1}^B \left( r_k(\mathbf{s}_{j,t}^{(k)},\mathbf{a}_{j,t}^{(k)})-r_k( \tilde{\mathbf{s}}_{j,t}^{(k)},\tilde{\mathbf{a}}_{j,t}^{(k)})\right)\right]
\end{split}
\end{equation}
where the action $\tilde{\mathbf{a}}_{i,t}^{(k)}\sim\pi_{{\theta_k}}^{Greedy}$ is from the greedy policy  $\pi_{{\theta_k}}^{Greedy}$, i.e. the action is sampled greedily, and $\tilde{\mathbf{s}}_{i,t}^{(k)}$ is the corresponding state. The second term of Equation \ref{eq:central-self-critic} is the gap of the rewards between the sampling and greedy approach, which is designed to centre the advantage term in the REINFORCE algorithm \cite{williams1992simple}. Using a central self-critic baseline accelerates the convergence rate compared to using an exponential moving average of the rewards.

Since the lowest layer of the hierarchy is a Markov Decision Process (MDP), the lowest-level policy is learned directly and provides latent variables for the higher layer. In other words, we use a bottom-up approach for learning the hierarchical policy and training the neural network.

\subsubsection{Layer-wise Policy Optimization}
Suppose we need to learn a $(K+1)$-layer hierarchical policy, which includes $\pi_{\theta_0}$,$\pi_{\theta_1}$,...,$\pi_{\theta_K}$. Each policy is represented by a GPN. In order to learn policy $\pi_{\theta_K}$, we first need to train all lower layers $\pi_{\theta_k}$ for $k=0,...,K-1$ and fix the weights of the neural networks. Then, for layer $k=0,...,K-1$, we sample $(\mathbf{s}_{t}^{(k)},\mathbf{a}_{t}^{(k)})$ based on $\pi_{\theta_k}$, and provide latent variable $\mathbf{h}_{t}^{(k+1)}$ for the next higher layer. Finally, we can learn the policy $\pi_{\theta_K}$ from $\mathbf{h}_{t}^{(K)}$. Algorithm \ref{alg:policyopt} provides detailed pseudo-code.

\begin{algorithm}
\caption{Layer-wise Policy Optimization}
\label{alg:policyopt}
\begin{algorithmic}[1]
\Procedure{Train}{training set $\mathcal{X}$, \# of training steps $M_0,M_1,...,M_K$, batch size $B$, learning rate $\alpha$, the number of layers $K$}
\State Initialize network parameters $\theta_k$ for $k\in\{0,...,K\}$
\For{$k=0$ to $K$}
\For{$m=1$ to $M_k$}
\State $\mathbf{X}_i\sim$Sample($\mathcal{X}$) for $i\in\{1,...,B\}$
\For{$j=0$ to $k-1$}
\State $\mathbf{a}_{i,t}^{(j)},\mathbf{h}_{i,t}^{(j+1)}\sim\pi_{\theta_j}(\cdot|\mathbf{s}_{i,t}^{(j)},\mathbf{h}_{i,t}^{(j)})$ 
\EndFor
\State $\mathbf{a}_{i,t}^{(k)}\sim\pi_{\theta_k}(\cdot|\mathbf{s}_{i,t}^{(k)},\mathbf{h}_{i,t}^{(k)})$
\State $\tilde{\mathbf{a}}_{i,t}^{(k)}\sim\pi_{\theta_k}^{Greedy}(\cdot|\tilde{\mathbf{s}}_{i,t}^{(k)},\mathbf{h}_{i,t}^{(k)})$
\State Compute $J(\theta_k),\nabla_{\theta_k}J(\theta_k)$
\State $\theta_k\leftarrow\theta_k+\alpha\nabla_{\theta_k}J(\theta_k)$
\EndFor
\EndFor
\State \Return $\pi_{\theta_0},\pi_{\theta_1},...,\pi_{\theta_K}$
\EndProcedure
\end{algorithmic}
\end{algorithm}

\section{GRAPH POINTER NETWORK}\label{sec:gpn}
\subsection{GPN Architecture}
\begin{figure}[ht]
\centering
\includegraphics[width=0.48\textwidth]{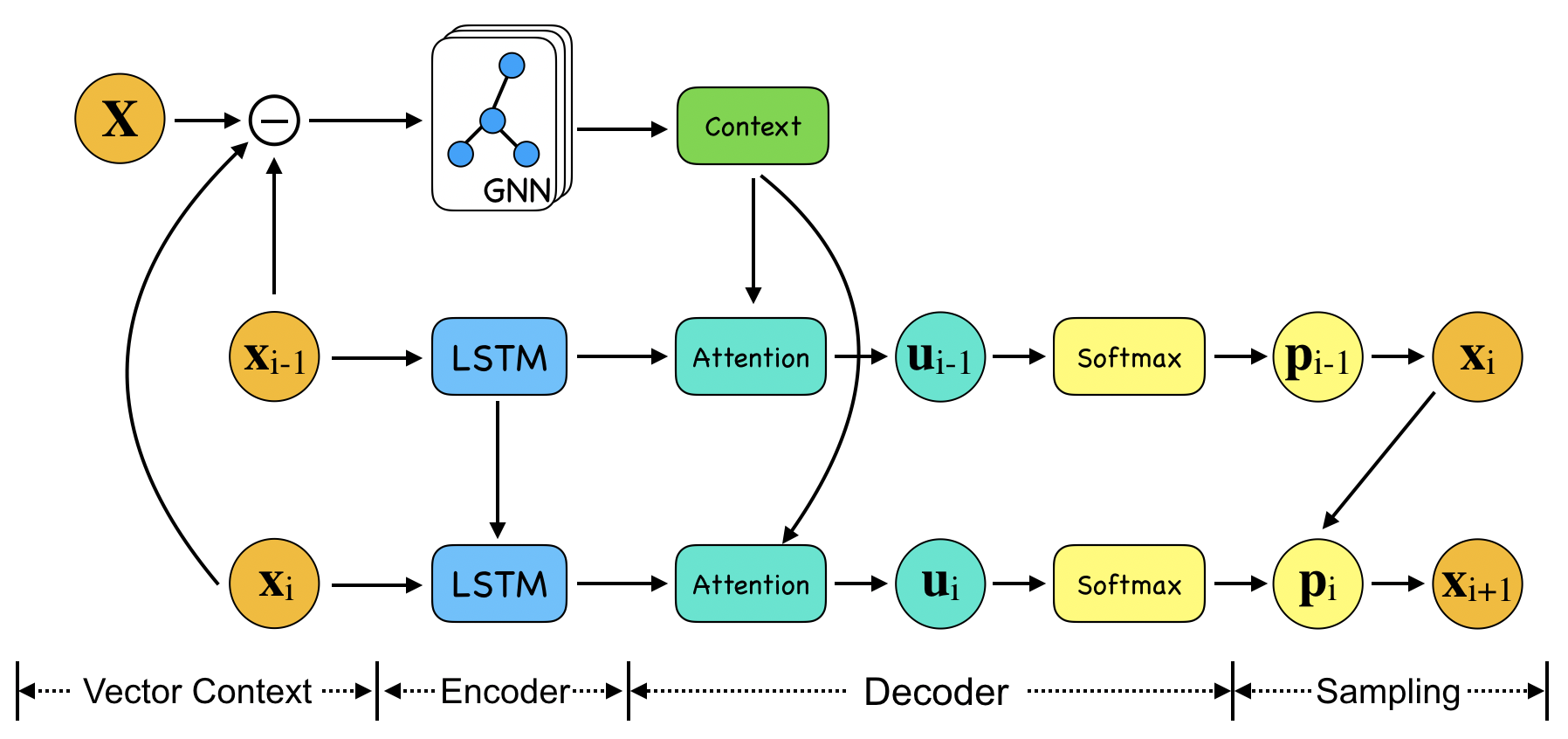}
\caption{\label{gpn} Architecture for Graph Pointer Network. The current city coordinate $\mathbf{x}_i$ (we denote  $\mathbf{x}_{\sigma(i)}$ as $\mathbf{x}_i$ for convenience) is encoded by the LSTM while $\bar{\mathbf{X}}=(\mathbf{X}-\mathbf{X}_i)$ is encoded as the vector context by a graph neural network. The encoded vectors are passed to the attention decoder, which outputs the pointer vector $\mathbf{u}_{i}$. The probability distribution over the next candidate city is $\mathbf{p}_{i}=\mbox{softmax}(\mathbf{u}_{i})$. The next visited city $\mathbf{x}_{i+1}$ is sampled from $\mathbf{p}_{i}$.}
\end{figure}

We propose a \textit{graph pointer network} (GPN) based on the pointer network \cite{bello2016neural} for approximately solving the TSP. The GPN architecture, which is shown in Figure \ref{gpn}, consists of an encoder and decoder component.

\paragraph{Encoder}
The encoder includes two parts: point encoder and graph encoder. For the point encoder, each city coordinate $\mathbf{x}_i$ is embedded into a higher dimensional vector $\tilde{\mathbf{x}}_i\in\mathbb{R}^d$, where $d$ is the hidden dimension. This linear transformation shares weights across all cities $\mathbf{x}_i$. The vector $\tilde{\mathbf{x}}_i$ for the current city $\mathbf{x}_i$ is then encoded by an LSTM. The hidden variable $\mathbf{x}^h_i$ of the LSTM is passed to both the decoder in the current step and the encoder in the next time step. For the graph encoder, we use graph embedding layers to encode all city coordinates $\mathbf{X}= [\mathbf{x}_1^{\top},...,\mathbf{x}_N^{\top}]^{\top}$, and pass it to the decoder. 


\paragraph{Graph Embedding Layer}
In TSP, the context information of a city node includes the neighbors' information of the city. In a GPN, context information is obtained by encoding all city coordinates $\mathbf{X}$ via a graph neural network (GNN) \cite{kipf2016semi,xu2018powerful}. Each layer of the GNN is expressed as:
\begin{equation}
\mathbf{x}_i^{l} = \gamma\mathbf{x}_i^{l-1}\Theta + (1-\gamma)\phi_{\theta}\left( \frac{1}{ |\mathcal{N}(i)|}\{\mathbf{x}_j^{l-1}\}_{j\in\mathcal{N}(i)\cup \{i\} }\right),
\end{equation}
where $\mathbf{x}_i^{l}\in\mathbb{R}^{d_{l}}$ is the $l$-th layer variable with $l\in\{1,...,L\}$, $\mathbf{x}_i^{0}=\mathbf{x}_i$, $\gamma$ is a trainable parameter which regularizes the eigenvalue of the weight matrix, $\Theta\in\mathbb{R}^{d_{l-1}\times d_{l}}$ is a trainable weight matrix, $\mathcal{N}(i)$ is the adjacency set of node $i$, and $\phi_\theta :\mathbb{R}^{d_{l-1}}\rightarrow\mathbb{R}^{d_{l}}$ is the aggregation function \cite{kipf2016semi}, which is represented by a neural network in this work. Furthermore, since we only consider symmetric TSP, the graph of the TSP is a complete graph. Therefore, the graph embedding layer is further expressed as:
\begin{equation}\mathbf{X}^{l} = \gamma\mathbf{X}^{l-1}\Theta + (1-\gamma)\Phi_{\theta}\left(\mathbf{X}^{l-1}/|\mathcal{N}(i)|  \right), 
\end{equation}
where $\mathbf{X}^{l}\in\mathbb{R}^{N\times d_{l}}$, and $\Phi_\theta :\mathbb{R}^{N\times d_{l-1}}\rightarrow\mathbb{R}^{N\times d_{l}}$ is the aggregation function.

\paragraph{Vector Context}
In previous work \cite{bello2016neural,kool2018attention}, the context is computed based on the 2D coordinates of all cities, i.e.  $\mathbf{X}\in\mathbb{R}^{N\times 2}$. We refer to this context as \textit{point context}. In contrast, instead of using coordinate features directly, in this work, we use the vectors pointing from the current city to all other cities as the context, which we refer to as a \textit{vector context}. For the current city $\mathbf{x}_i$, suppose $\mathbf{X}_i=[\mathbf{x}_i^{\top},...,\mathbf{x}_i^{\top}]^{\top}\in\mathbb{R}^{N\times 2}$ is a matrix with identical rows $\mathbf{x}_i$. We define $\bar{\mathbf{X}}_i = \mathbf{X}-\mathbf{X}_i$ as the vector context. The $j$-th row of $\bar{\mathbf{X}}_i$ is a vector pointing from node $i$ to node $j$. Then $\bar{\mathbf{X}}_i$ is passed into the graph embedding layers. A graph embedding layer is rewritten as $\bar{\mathbf{X}}_i^{l} = 
\gamma\bar{\mathbf{X}}_i^{l-1}\Theta + (1-\gamma)\Phi_{\theta}\left( \bar{\mathbf{X}}_i^{l-1}/|\mathcal{N}(i)|  \right)$. In practice, the GPN using the vector context yields more transferable representations, which allows the model to perform well on larger-scale TSP.

\paragraph{Decoder}
The decoder is based on an attention mechanism and outputs the pointer vector $\mathbf{u}_i$, which is then passed to a softmax layer to generate a distribution over the next candidate cities. Similar to pointer networks \cite{bello2016neural}, the attention mechanism and pointer vector $\mathbf{u}_i$ is defined as:
\begin{equation}
\mathbf{u}_i^{(j)}=
\begin{cases}
v^{\top}\cdot\mathrm{tanh}(W_r r_j+ W_q q) & \mbox{if}~ j\neq\sigma(k), \forall k<j,\\
-\infty &  \mbox{otherwise,}\\
\end{cases}
\end{equation}
where $\mathbf{u}_i^{(j)}$ is the $j$-th entry of the vector $\mathbf{u}_i$, $W_{r}$ and  $W_{q}$ are trainable matrices, $q$ is a query vector from the hidden variable of the LSTM, and $r_i$ is a reference vector containing the information of the context of all cities. Precisely, we use the hidden variable $\mathbf{x}^h_i$ from the point encoder as the query vector $q$, and use the context $\mathbf{X}^L$ from the graph embedding layer as the reference, i.e. $q=\mathbf{x}^h_i$ and $r_j=\mathbf{X}^L_j$.

The distribution policy over all candidate cities is given by:
\begin{equation}
\pi_{\theta}(\mathbf{a}_{i}|\mathbf{s}_{i})=\mathbf{p}_i=\mbox{softmax}(\mathbf{u}_i)
\end{equation}
We predict the next visited city $\mathbf{a}_i=\mathbf{x}_{\sigma(i+1)}$,
by sampling or choosing greedily from the policy $\pi_{\theta}(\mathbf{a}_{i}|\mathbf{s}_{i})$.

\subsection{Hierarchical GPN Architecture}
\begin{figure}[ht]
\centering
\includegraphics[width=0.45\textwidth]{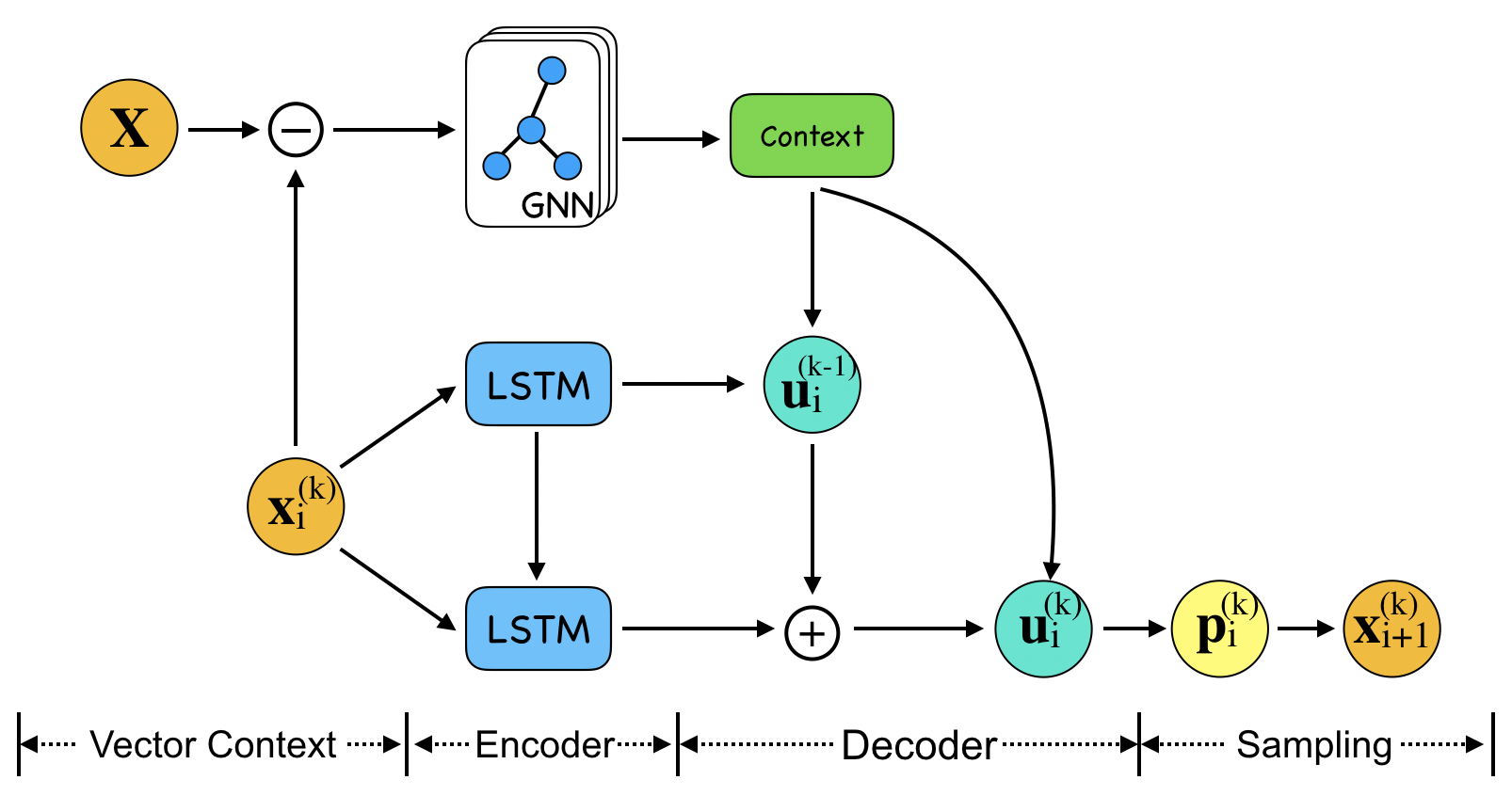}
\caption{\label{Hpointer} A two layer hierarchical architecture of GPN. The pointer vectors of the two layers are added together to predict the next candidate city. The pointer vector of the lower layer provides a prior for the higher layer.}
\end{figure}

In this section, we use the proposed GPN to design a hierarchical architecture. The architecture of a two-layer hierarchical GPN (HGPN) is illustrated in Figure \ref{Hpointer}. In contrast to a single-layer GPN, the coordinate $\mathbf{x}_i^{(k)}$ at $k$-th layer is first passed as input to a lower-level neural network and the network outputs a pointer vector $\mathbf{u}_{i}^{(k-1)}$. Then, $\mathbf{u}_{i}^{(k-1)}$ is added to the pointer vector $\mathbf{u}_{i}^{(k)}$ of a higher layer, i.e. $\mathbf{p}_i^{(k)}=\mbox{softmax}\left(\mathbf{u}_i^{(k)}+\alpha\mathbf{u}_{i}^{(k-1)}\right)$, where $\alpha$ is a trainable parameter. This plays an important role since $\mathbf{u}_{i}^{(k-1)}$ contains lower layer information which provides a prior distribution over the output cities. The output $\mathbf{x}_{i+1}^{(k)}$ is then sampled from $\pi_{\theta}(\cdot|\mathbf{s}_{i}^{(k)},\mathbf{h}_{i}^{(k)})=\mathbf{p}_i^{(k)}$, where $\mathbf{h}_{i}^{(k)}=\mathbf{u}_{i}^{(k-1)}$ is the latent variable from the lower layer.

\section{EXPERIMENTS}
\label{sec:experiments}
In our experiments, we use $L=3$ graph embedding layers to encode the context in the GPN. The aggregation function used is a single layer fully connected neural network. The graph embedding layer is expressed as 
\begin{equation}
\mathbf{X}^{l} = \gamma\mathbf{X}^{l-1}\Theta + (1-\gamma)g( \mathbf{X}^{l-1}W/|\mathcal{N}(i)|+b),
\end{equation}
where $g(\cdot)$ is the ReLU activation function, $W\in\mathbb{R}^{d_{l-1}\times d_a}$ and $b\in\mathbb{R}^{N\times d_a}$  are trainable weights and biases with $d_l=d_a=128$ for $l=1,2,3$. 
We use point context for small-scale problems such as TSP20/50 and vector context for larger-scale problems such as TSP500. 
The training data is generated randomly from a $[0,1]^2$ uniform distribution. In each epoch, the training data is generated on the fly. The central self-critic baseline is used during RL training. Unless otherwise specified, the following experiments use the hyperparameters shown in Table \ref{t-param}.
\begin{table}
\centering
\caption{Hyperparameters used for training}
\label{t-param}
\begin{tabular}{lc|lc}
\hline
Parameter & Value & Parameter & Value\\ \hline
Epoch & 100 & Optimizer  &  Adam \\
Batch size & 512 & Learning rate  &  1e-3 \\
Training steps (per epoch) & 2500 &  Learning rate decay & 0.96 \\
\hline
\end{tabular}
\end{table}

\paragraph{OR-Tools Setting}
We use OR-Tools \cite{google_2016} as one of the baselines to compare with our result. To compare with larger-scale TSP instances, the Savings and Christofides algorithms are selected as first solution strategies in OR-Tools. The search time limit for each TSP instance is set to 5 seconds. We choose Guided Local Search as the metaheuristic when running OR-Tools. 
For TSP with Time Windows (TSPTW), the Savings algorithm is picked as the first solution strategy in OR-Tools. We use the default setting for its search limits and metaheuristics.

\subsection{Experiments for small-scale TSP}

\begin{figure}[ht]
\centering
\includegraphics[width=0.45\textwidth]{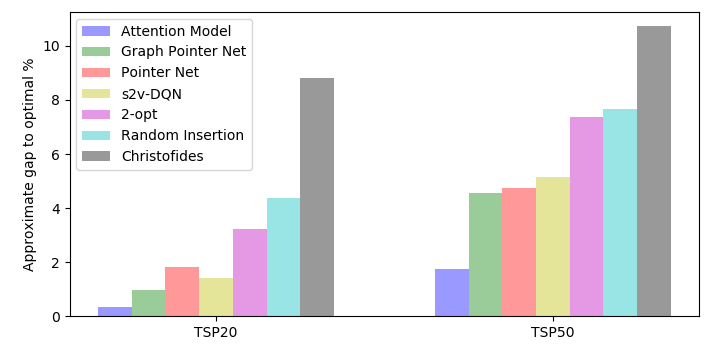}
\caption{\label{compare} Comparison of TSP20/50 results: Attention Model, s2v-DQN, Pointer Net, 2-opt, random insertion and Christofides. The y-axis is the approximate gap to the optimal solutions.}
\end{figure}

We train our GPN model with TSP20 and TSP50 instances. The training time of each epoch is 7 minutes for TSP20 and 30 minutes for TSP50 using one NVIDIA Tesla P100 GPU. We compare the performance of our model on small-scale TSP with previous work, such as the Attention Model \cite{kool2018attention}, s2v-DQN \cite{khalil2017learning}, the Pointer Network \cite{bello2016neural}, and other heuristics, e.g. 2-opt heuristics, Christofides algorithm and random insertion. 
The results are shown in Figure \ref{compare}, which compares the approximate gap to the optimal solution. A smaller gap indicates a better result. The optimal solutions are obtained from the LKH algorithm. 
We observe that for small-scale TSP instances, the GPN outperforms the Pointer Network, which demonstrates the usefulness of the graph embedding, but yields worse approximations than the Attention Model. 

%
%

\subsection{Experiments for larger-scale TSP}
\begin{table*}
\centering
\caption{Comparison for larger-scale TSP. Each result is obtained by running on 1000 random TSP instances. Tour Len refers to average tour length. Time refers to total running time (sec) of 1000 instances.}
\label{tab:large_scale}
\begin{tabular}{l|cc|cc|cc|cc} 
\hline
\hline
& \multicolumn{2}{c|}{TSP 250} & \multicolumn{2}{c|}{TSP 500} & \multicolumn{2}{c|}{TSP 750} & \multicolumn{2}{c}{TSP 1000} 
\\ 
Method & Tour Len. & Time & Tour Len. & Time
& Tour Len. & Time & Tour Len. & Time\\ 
\hline\hline
LKH & 11.893 & 9792s & 16.542 & 23070s & 20.129 & 36840s & 23.130 & 50680s \\
Concorde & $11.89$ & 1894s & $16.55$ & 13902s & $20.10$ &  32993s& $23.11$& 47804s\\
Nearest Neighbor & 14.928 & 25s & 20.791 & 60s & 25.219 & 115s & 28.973 & 136s\\
2-opt & 13.253 & 303s & 18.600 & 1363s & 22.668 & 3296s & 26.111 & 6153s\\
Farthest Insertion & 13.026 & 33s & 18.288 & 160s & 22.342 & 454s & 25.741& 945s\\
OR-Tools (Savings) & 12.652 & 5000s & 17.653 & 5000s & 22.933 & 5000s & 28.332 & 5000s\\
OR-Tools (Christofides) & 12.289 & 5000s & 17.449 & 5000s & 22.395 & 5000s & 26.477 & 5000s\\
\hline
s2v-DQN & 13.079 & 476s & 18.428 & 1508s & 22.550 & 3182s & 26.046 & 5600s\\
Pointer Net & 14.249 & 29s & 21.409 & 280s & 27.382 & 782s & 32.714 & 3133s \\
Attention Model & 14.032 & 2s & 24.789 & 14s & 28.281 & 42s & 34.055 & 136s\\
\textbf{GPN (ours)} & 13.679 &  32s & 19.605 & 111s & 24.337 & 232s & 28.471 & 393s\\
\textbf{GPN+2opt (ours)} &12.942 & 214s & 18.358 & 974s & 22.541 & 2278s & 26.129 & 4410s\\
\hline
\end{tabular}
\end{table*}

In real world applications, most practical TSP instances have hundreds or thousands of nodes, and the optimal solution is not efficiently computable. We find that the proposed GPN model generalizes well from small-scale TSP problems to larger-scale problems. The generalization capacity increases by an order of magnitude.

In Table \ref{tab:large_scale}, we train a GPN model with vector context on TSP50 data with 10 epochs, and use this model to predict the routes on TSP250/500/750/1000. Furthermore, we use a local search algorithm 2-opt \cite{aarts2003local} to improve our results after prediction. The Pointer Network (PN) \cite{bello2016neural}, s2v-DQN \cite{khalil2017learning} and Attention Model (AM) \cite{kool2018attention} are also trained with TSP50 data, and we check the transferability of these models to larger-scale problem as well. Results are averaged over 1000 TSP instances. Due to memory constraints, we set the batch size $B=50$ during inference for all models. The results are also compared with LKH, nearest neighbor, 2-opt, farthest insertion and Google OR-Tools \cite{google_2016}.

\begin{figure}[ht]
\centering
\subfigure[TPS250 (GPN+2opt)]{
\includegraphics[width=0.20\textwidth]{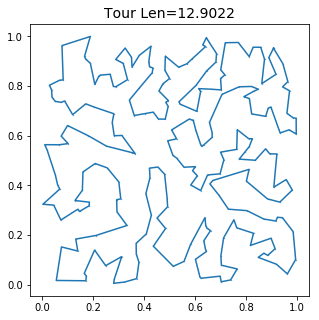}}
\subfigure[TPS500 (GPN+2opt)]{
\includegraphics[width=0.20\textwidth]{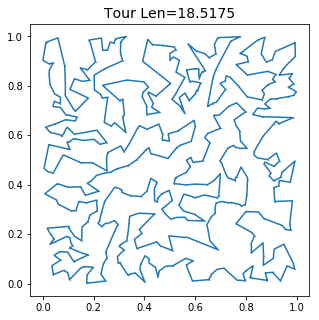}}
\subfigure[TPS750 (GPN+2opt)]{
\includegraphics[width=0.20\textwidth]{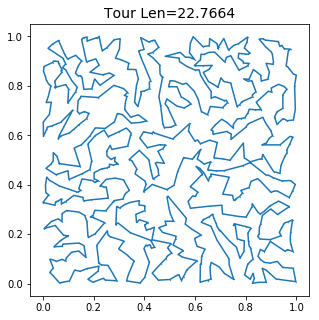}}
\subfigure[TPS1000 (GPN+2opt)]{
\includegraphics[width=0.20\textwidth]{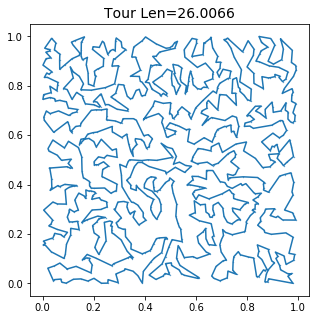}}
\caption{Sample tours for TSP250/500/750/1000. Approximate solutions of larger-scale TSP predicted by GPN and 2-opt heuristics.}
\label{fig:sample}
\end{figure}

Table \ref{tab:large_scale} shows that our GPN model outperforms PN and AM when we train with TSP50 instances and generalize to larger-scale problems. With local search added, the GPN+2opt has similar tour length to s2v-DQN, but saves $\approx 20\%$ running time. Compared with the 2-opt heuristic, the GPN+2opt uses $\approx 25\%$ less running time, which means the GPN model can be treated as a good initialization method. The GPN+2opt also outperforms OR-Tools on TSP1000. On Table \ref{tab:large_scale}, GPN does not outperform the state-of-the-art TSP solver, e.g. LKH and Farthest Insertion. However, it still has the potential to be an effective initialization method, since the GPN shows good generalization capabilities and can solve TSP instances in parallel. Some sample tours are shown in Figure \ref{fig:sample}.

As aforementioned, the generalization capacity of the GPN model is roughly an order of magnitude larger than the size of the instances the model is trained on. More specifically, we train the GPN models on TSP20/50/100 and use these models to predict on TSP500/1000. The results are shown in Table \ref{tab:large_general}, which demonstrates that the results improve if we increase the size of the TSP instances used for training.

\begin{table}
\centering
\caption{Comparison for larger-scale TSP. The GPNs are trained with different size of TSP instances. Each result is obtained by running on 1000 random TSP instances. Tour Len refers to average tour length. Time refers to total running time (sec) of 1000 instances.}
\label{tab:large_general}
\begin{tabular}{l|cc|cc} 
\hline
\hline
& \multicolumn{2}{c|}{TSP 500} & \multicolumn{2}{c}{TSP 1000} 
\\ 
Model & Tour Len. & Time & Tour Len. & Time\\ 
\hline
GPN (TSP20)& 22.320 & 107s & 33.649 & 391s\\
GPN (TSP50) &  19.605 & 111s & 28.471 & 393s\\
GPN (TSP100) &\bf{19.527} & 109s & \bf{28.036} & 408s\\
\hline
\end{tabular}
\end{table}

\subsection{Experiments for TSP with time window}
Finally, we consider a well known constrained TSP problem, the TSP with Time Windows (TSPTW). In TSPTW, each node $i$ has its own service time interval $[e_i,l_i]$, where $e_i$ is the entering time and $l_i$ is the leaving time. A city cannot be visited after its leaving time. If the node is visited earlier than the entering time, the salesman must wait until the service begins, namely until the entering time. In this experiment, we consider the following formalization of the TSP with Time Windows problem:
\begin{equation}\label{tsptw6}
\begin{array}{cll}
\min\limits_{\sigma}  & \sum\limits_{i=1}^{N} c_i & \\
 \mbox{s.t.} &  c_{i+1}-c_{i}\geq  \|\mathbf{x}_{\sigma(i+1)}-\mathbf{x}_{\sigma(i)}\|_2, & i\in\{1,...,N-1\},\\
 & e_i\leq c_i \leq l_i & i\in\{1,...,N\},
\end{array}
\end{equation}
where $c_i$ is the time cost for the $i$-th city. In this problem, a feasible solution does not always exist. To ensure the existence of training and test data, we first generate TSP20 instances from a $[0,1]^2$ uniform distribution. Then using 2-opt local search on the generated instances, we solve the approximate solutions $\tilde{c}_i$ for $i\in\{1,...,N\}$. We set $e_i=\max\{\tilde{c}_i-\tilde{e}_i,0\}$ and $l_i =\tilde{c}_i+\tilde{l}_i$, where $\tilde{e}_i\sim\mathrm{Uniform}(0,2)$ and $\tilde{l}_i\sim\mathrm{Uniform}(0,2)+1$. Therefore, $e_i\leq \tilde{c}_i\leq l_i$, which means that a feasible solutions in the training and test data always exist. The dataset is obtained by shuffling all cities in the instances above. The exponential moving average critic baseline \cite{kool2018attention} is used during RL training.

In the experiments for TSP with Time Windows (TSPTW), we construct a two-layer hierarchical GPN (HGPN). First we define:
\begin{equation}
\rho(c,l)\coloneqq\sum_{i=1}^{N}\max\{l_i-c_i, 0\},
\end{equation}
as the penalty if the arriving time exceeds the leaving time, where $l_i$ is the leaving time and $c_i$ is the arriving time. Then the reward function of the lower layer is the penalty of violating the leaving time constraints $r_1 = \beta*\rho(c,l)$, where $\beta$ is the penalty factor. The reward of the higher layer is the total time cost of TSPTW plus the penalty:
\begin{equation}
r_2 = \sum_{i=1}^{N}c_i+\beta*\rho(c,l).
\end{equation}
For the inference phase, we use $\rho(c,l)$ to measure accuracy, i.e. the number of instances that are solved successfully. For any of the instances, if $\rho(c,l)>0$, then there exists at least one city such that the arriving time exceeds the leaving time, which indicates that the solution is infeasible.

The lower layer is trained with $1$ epoch of TSPTW20 data, and the higher layer is trained with $19$ epochs. For TSPTW data, each of the nodes $\mathbf{x}_i$ is a tuple $(x_i,y_i,e_i,l_i)$, where $(x_i,y_i)$ is a 2-D coordinate and $e_i, l_i$ are the entering and leaving time. We average results over $10000$ problem instances to compare our results with OR-Tools and Ant Colony Optimization (ACO) algorithm
\cite{cheng2007modified}. 

\begin{figure}[ht]
\centering
\subfigure{
\includegraphics[width=0.21\textwidth]{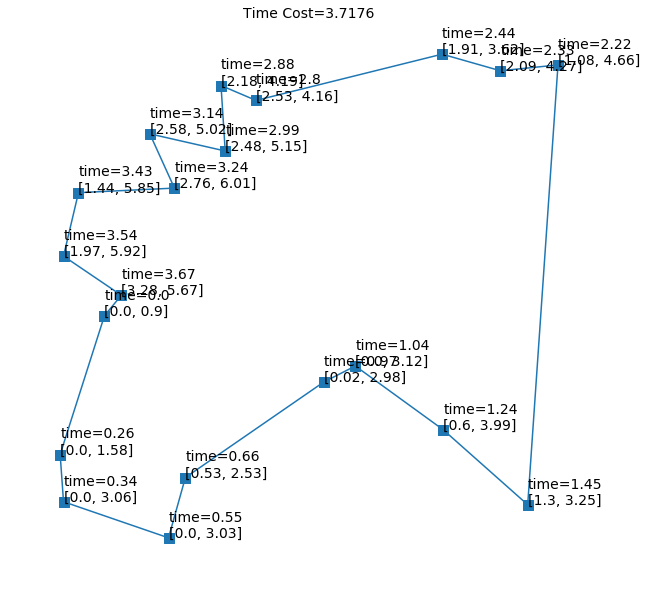}}
\subfigure{
\includegraphics[width=0.21\textwidth]{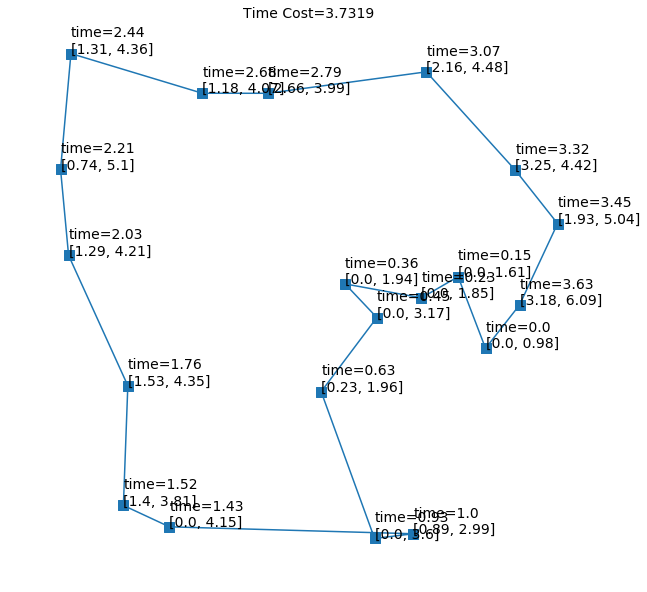}}
\caption{Sample tours for TSPTW20. For the text on each node, the first line is the arriving time, and the second line is the time window.}
\label{fig:tsptw}
\end{figure}

At prediction time, we use both the greedy and sampling method. The result is improved by sampling 100 or 500 times. Table \ref{t-tsptw} demonstrates that our HGPN framework outperforms all other baselines on TSPTW including the single-layer GPN. 
Even though all instances have feasible solutions based on our training setup, sometimes the algorithms will fail to find a feasible solution. To capture this, we use the percentage of feasible solutions as an evaluation metric. The HGPN achieves a much higher percentage of feasible solutions compared to the baselines. Some sample tours are shown in Figure \ref{fig:tsptw}.
\begin{table}
\centering
\caption{Results for TSPTW20. Cost: objective of TSPTW. Time: the running time of the algorithms. Feasible \%: the percentage of instances that are predicted to have feasible solutions by the algorithm.}
\label{t-tsptw}
\begin{tabular}{l|ccc}
\hline
Method & Cost &  Time & Feasible \% \\ \hline
OR-Tools (Savings) & 4.045 & 121s & 72.06\% \\
ACO  & 4.655 & 204s  &  62.10\%\\
\hline
GPN-greedy  & 4.209 & 1s & 99.87\% \\
HGPN-greedy  & 4.178 & 1s & 99.88\% \\
HGPN-sampling-100  & 4.013 & 99s & 100\%\\
HGPN-sampling-500  & \bf{3.991} & 494s & 100\%\\
\hline
\end{tabular}
\end{table}

\subsection{Real World TSP Instances}
We have evaluated our model on the real world TSPLIB dataset using instances which have less than 1500 nodes. We report the average gap between our result and the best solution, which is shown in Table \ref{tab:tsplib}.
\begin{table}
\centering
\caption{Evaluation on real world TSPLIB dataset.\newline}
\label{tab:tsplib}
\begin{tabular}{l|cc}
\hline
Method & Concorde &  GPN+2opt  \\ \hline
Optimality Gap & $0.13\pm0.6\%$ & $9.35\pm3.45\%$ \\
Running Time &1377s & 200s\\
\hline
\end{tabular}
\end{table}\\

\section{DISCUSSION}
\subsection{Generalization}
\begin{figure}[H]
\centering
\includegraphics[width=0.30\textwidth]{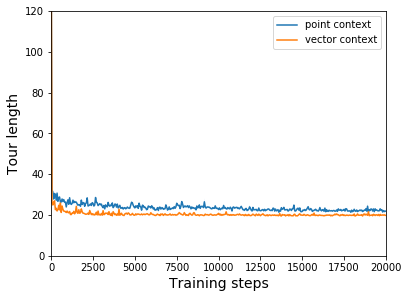}
\caption{Validation curves of GPN on TSP500. The GPN model is trained with TSP50 and generalizes on TSP500.}
\label{generalization}
\end{figure}

\paragraph{Vector Context}
In our GPN model, we use vector contex before encoding. The vector context is helpful to obtain pairwise information between cities. Therefore, in each step, our model knows the relative position between the current city and all others, which contributes to good generalization. In the experiments, the GPN with vector context performs better than GPN with point context on larger-scale TSP, which is illustrated in Figure \ref{generalization}.

\paragraph{Early Stopping} 
In order to generalize well to larger-scale instances and avoid overfitting on small-scale problems, we use early stopping during training and solve larger-scale TSP with models trained for 10 epochs. The comparison results between performance on various levels of early stopping is shown in Table \ref{tab:early_stop}. Based on the performance, we still train PN, s2v-DQN, AM for 100 epochs.
\begin{table}
\centering
\caption{Tour length results obtained by GPN on different epochs. We train GPN with TSP50 instances and predict TSP500/1000. The bold result is the shortest tour length.}
\label{tab:early_stop}
\begin{tabular}{l|cccc} 
\hline
Epoch & 1 & 5 & \bf{10} & 100 \\
\hline TSP500 & 20.26 & 19.69 & \bf{19.58} & 20.19\\
TSP1000 & 29.23 & 28.52 & \bf{28.48} & 29.28\\
\hline
\end{tabular}
\end{table}

\paragraph{Clip Range} In our model, we clip the range of pointer vector $\mathbf{u}$ to $[-C, C]$. Instead of $C=10$, which is used in previous work \cite{bello2016neural}, we choose $C=100$ to make a better exploration-exploitation tradeoff.

\subsection{Hierarchical Architecture}
\begin{figure}[H]
\centering
\includegraphics[width=0.30\textwidth]{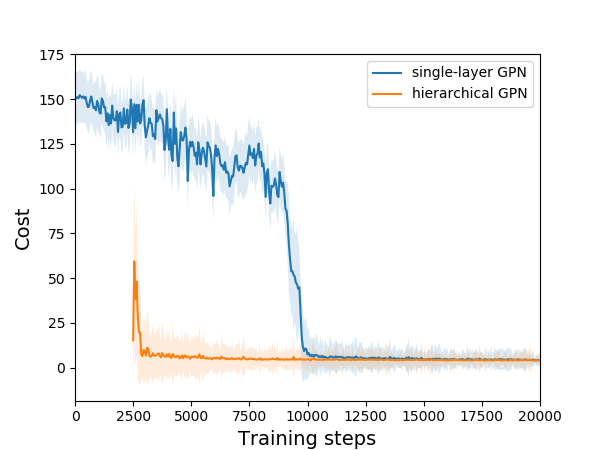}
\caption{Validation curves of HGPN and single-layer GPN on TSPTW20 during training. The orange curve shows only the higher layer of HGPN and begins at the 2nd epoch. The 1st epoch is used for training lower layer.} 
\label{hier_vs_single}
\end{figure}

In TSPTW problem, the hierarchical GPN (HGPN) performs better than single-layer GPN. The training curves of HGPN and single-layer GPN are shown in Figure \ref{hier_vs_single}. For single-layer GPN, the reward function includes both the penalty and the objective of TSPTW, which leads to unstable training on the early stage as shown by the blue curve in Figure \ref{hier_vs_single}. In contrast, we train the lower layer of HGPN to minimize the penalty term, which is simple to learn and converges quickly within one epoch. Then, the lower layer provides a prior distribution of possible feasible solutions for the higher layer. Given the latent information of feasible solutions, the higher layer of HGPN converges quicker than single-layer GPN and yields better solutions. 

\section{CONCLUSION}\label{sec:conclusion}
In this work, we propose a Graph Pointer Network (GPN) framework which efficiently solves larger-scale TSP by using graph embedding layers. Training a hierarchical RL model allows our approach to additionally tackle constrained combinatorial optimization problems such as the TSP with time windows. Our experimental results demonstrate that the GPN generalizes well from small-scale to larger-scale problems, outperforming previous RL methods. We make our data, models, and code publicly available \cite{ma2019tsp}.

\bibliography{ecai}
\end{document}